# DeepStory: Video Story QA by Deep Embedded Memory Networks

**Kyung-Min Kim**[1,2], **Min-Oh Heo**[1], **Seong-Ho Choi**[1], and **Byoung-Tak Zhang**[1,2]
[1]School of Computer Science and Engineering, Seoul National University
[2]Surromind Robotics
{kmkim, moheo, shchoi, btzhang}@bi.snu.ac.kr

## Abstract

Question-answering (QA) on video contents is a significant challenge for achieving human-level intelligence as it involves both vision and language in real-world settings. Here we demonstrate the possibility of an AI agent performing video story QA by learning from a large amount of cartoon videos. We develop a video-story learning model, i.e. Deep Embedded Memory Networks (DEMN), to reconstruct stories from a joint scene-dialogue video stream using a latent embedding space of observed data. The video stories are stored in a long-term memory component. For a given question, an LSTM-based attention model uses the long-term memory to recall the best question-story-answer triplet by focusing on specific words containing key information. We trained the DEMN on a novel QA dataset of children's cartoon video series, *Pororo*. The dataset contains 16,066 scene-dialogue pairs of 20.5-hour videos, 27,328 fine-grained sentences for scene description, and 8,913 story-related QA pairs. Our experimental results show that the DEMN outperforms other QA models. This is mainly due to 1) the reconstruction of video stories in a scene-dialogue combined form that utilize the latent embedding and 2) attention. DEMN also achieved state-of-the-art results on the MovieQA benchmark.

## 1 Introduction

The question-answering (QA) problem is an important research theme in artificial intelligence, and many computational models have been proposed during the past few decades. Most traditional methods focused on knowledge representation and reasoning based on natural language processing with many hand-crafted syntactic and semantic features [Abney *et al*., 2000; Hovy *et al*., 2001]. Recently, deep learning methods have started to outperform traditional methods on text domain using convolutional neural networks [Yu *et al.,* 2014], long short-term memory [Wang *et al*., 2015] and attention based deep models [Tan *et al*., 2016]. These methods are different from previous approaches in that they do not require any feature engineering and exploit a large amount of training data. The performance improvements have continuously extended to image QA tasks [Fukui *et al*., 2016; Kim *et al*., 2016]. However, the results in the video domain so far have lagged compared to that in the text or image settings. There are still very few methods and datasets to address the video story QA. [Kim *et al*., 2015] used a probablistic concept graph. [Tapaswi *et al*., 2016] built two story learning models separately from scenes and dialogues of videos and fused the final answer predictions of each model, i.e. averaging the answer prediction scores. This late fusion sometimes led to performance degradation because understanding video stories requires both scene and dialogue information together, not separately.

At this point, this paper provides two contributions to the video story QA problem. First, we construct a novel and large-scale video story QA dataset-PororoQA from children's popular cartoon videos series '*Pororo*'. PororoQA has two properties that make it suitable as a test bed for video story QA. 1) Due to the characteristics of cartoon videos series, it has simple, clear but a coherent story structure and a small environment compared to other videos like dramas or movies. 2) The dataset provides high-quality scene descriptions to allow high-level video analysis. The new dataset consists of 16,066 video scene-dialogue pairs created from the videos of 20.5 hours in total length, 27,328 fine-grained descriptive sentences for scene descriptions and 8,913 multiple choice questions about the video stories. Each question is coupled with a set of five possible answers; one correct and four incorrect answers provided by human annotators. We plan to release the dataset to the community.

Second, we propose a video story learning model - Deep Embedded Memory Networks (DEMN). DEMN reconstructs stories from a joint stream of video scene-dialogue by combining scenes and dialogues in sentence forms using a latent embedding space. The video stories are stored in the long-term memory component that can be read and written to [Ha *et al*., 2015; Weston *et al*., 2015]. For a given QA pair, a word level attention-based LSTM evaluates the best answer by creating question-story-answer triplets using long-term memory and focusing on specific keywords. These processes pass through three modules (video story understanding module, story selection module, answer selection module) and, they are learned in a supervised learning setting.

We test our model on two different datasets – PororoQA and MovieQA and compare the results with various story QA models including human, VQA models, memory networks.

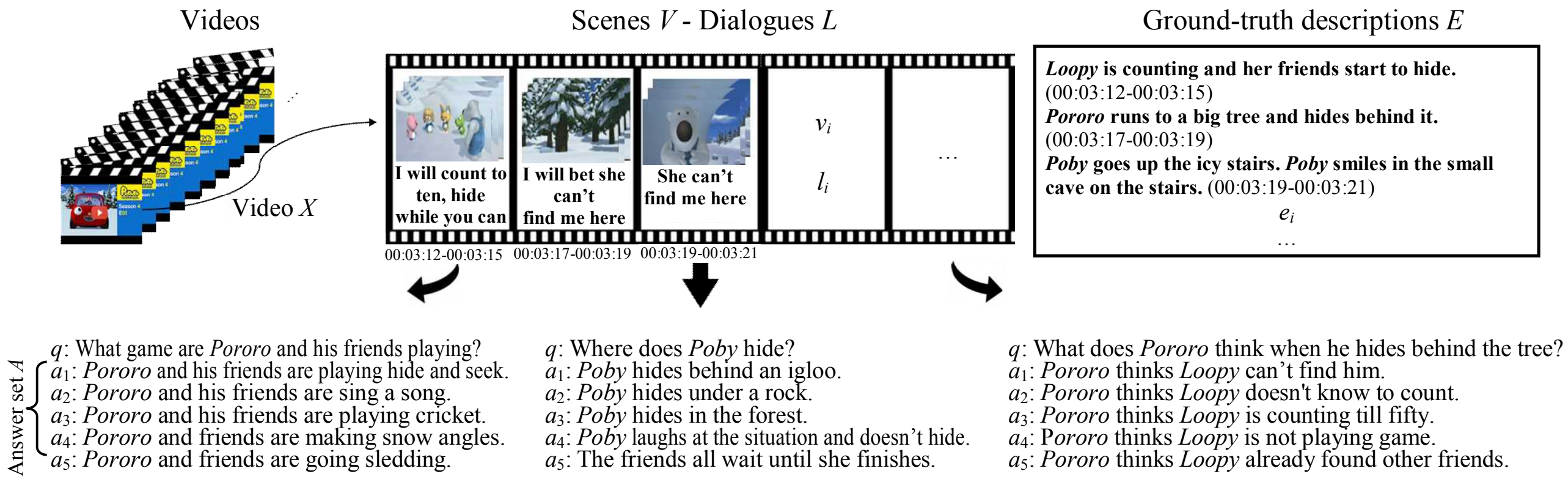

Figure 1: Overview of the video story QA task.

The experimental results show that the DEMN achieves state-of-the-art performances on both datasets by 1) the reconstruction of video stories in a scene-dialogue combined form using the latent embedding and 2) the attention.

## 2 Video Story QA

### 2.1 Task Statement

Figure 1 shows the overview of a video story QA task. We regard video $X$, e.g. an episode of a cartoon video series, as a compositional material consisting of a set of video scenes $V=\{v_i\}_{1...|V|}$, and a set of dialogues $L=\{l_i\}_{1...|L|}$, where $v_i$ is a sequence of image frames (not single). $l_i$ is a natural language sentence containing a dialogue. In our work, $v_i$ is paired with $l_i$, i.e. $X=\{(v_i,l_i)\}_{1...|X|}$, and $|X|=|V|=|L|$. Also, we assume that there is a set of externally available sources $E$ which is not included in the video itself but available for the video story QA task. For an external information source, we use a set of ground-truth scene descriptions $E$ in which a description $e_i$ is one or multiple sentences describing the scene $v_i$ of video $X$. Considering that most videos in the real environment do not have fine-grained annotations, we let the QA model learn the relationships between the ground-truth descriptions and scenes of videos during training and retrieve the relevant descriptions $\hat{e}$ from $E$ at test time for newly observed scene.

We also formulate a video story QA task as a multi-choice QA task. Each video $X$ has story-related questions and an answer sentence set for each question. Let $q$ be a question about a story of a video $X$ and, $A$ be a hypothesis answer sentence set for $q$. $A$ consists of $k$ multiple choice answer sentences $A=\{a_r\}_{1...k}$ (in our work, $k$=5). Thus, the QA model should choose a correct answer sentence among the set of $k$ possible answer sentences given a question $q$ and a video $X$. In other words, given a scoring function $F(X, q, a_r)$, our goal is to pick the correct answer sentence $a^*$ that maximize $F$:

$$a^* = \arg\max_{a_r} F(X, q, a_r) \quad (1)$$

### 2.2 Related Datasets for Video Story QA

From the extensive literature review, MovieQA is the most similar public dataset to the video story QA task [Tapaswi *et al.*, 2016]. This dataset contains 14K QA pairs about 408 movie stories and provides various information sources including video clips and descriptive sentences. However, to be more suitable as a testbed for the video story QA, certain points of MovieQA should be considered:

- Scenes of videos are not always provided for the QA tasks (6,462 questions can be answered using the scenes); most questions are answered only using the linguistic information sources such as dialogues, script, and plot.
- All questions are created from the plot synopsis of *Wikipedia* without watching movies.
- All movies have different characters and backgrounds, and complex and distant story structures that make optimization difficult.
- The descriptive sentences in MovieQA often contain contextual information not available within the provided video clip [Li *et al.*, 2016], i.e. low cohesion between the scenes and descriptive sentences.

For these reasons, we created a new benchmark that 1) allows high-level video analysis with high-quality, large amounts of descriptive sentences, and 2) have simple, clear but a coherent story structure.

### 2.3 PororoQA Dataset

Because of its simple nature, cartoon images were used for exploring the high-level reasoning required to solve imageQA [Zitnick *et al.*, 2013; Agrawal *et al.*, 2015]. Similar to cartoon images, cartoon videos have a simple story structure and a small environment compared to other videos such as movies and dramas. In particular, cartoon series for kids have the properties that similar events are repeated, and the number of characters and background is small. We use a famous cartoon video series for children called '*Pororo*', which consists of 171 episodes. Each episode has a different story of 7.2 minutes average length, and the amount of total running time is 20.5 hours. There are ten main characters in the entire video. The size of vocabulary is about 4,000.

**Scene & dialogue pair construction:** We extracted scenes by segmenting the videos based on the start/end times of speech of all the character in the videos and made 16,066 scene-dialogue pairs from the whole video. Note that the

Table 1: Polling results comparing the descriptions from MovieQA and PororoQA datasets.

| | MovieQA | PororoQA |
|---|---|---|
| Q1: Sentence only describes the visual information that can be obtained in the video. | 0.46 | 0.75 |
| Q2: Sentence precisely describes the scene without missing information. | 0.40 | 0.71 |

**Please read carefully. Your work will be rejected if you don't follow the guide lines.**

1. After selecting an episode and watching a Youtube video, please make story-related QAs in English. Please keep in mind that be sure to watch a video before making any data.
2. Please select a scene-subtitle pair that is most likely to match with your question and write down QAs. All questions should be localized in the video contents.
3. Please provide a correct answer to a question that most people would agree on and four wrong answers for deception.
4. Your answer should be a complete sentence with correct grammar. The minimum number of words in a sentence is four.
   - "lunch" (x) -> "*Pororo* and *Crong* are having a lunch" (o)
5. Please avoid vague terms.
   - "When did *Pororo* go there?" (x) -> "When did *Pororo* go *Loopy*'s house?" (o)
   - "What is *Pororo* doing?" (x) -> "What is *Pororo* doing when *Crong* is crying in the house?" (o)
6. Please avoid completely unrelated questions.
   - "How old is the earth" (x)
7. Please avoid image-specific (and not story-related) questions.
   - "How many trees are in the video?" (x)
8. Please avoid creating duplicate questions in an episode.
9. Please use character names as follows.

Figure 2: The instructions shown to the AMT QA creators.

subtitles of the videos were used to make the dialogues; thus, the dialogues do not contain speaker information. A scene has 34 image frames on average.

**Descriptive sentences collection:** The descriptive sentences and QAs were collected through our website. We made all the videos and the scene-dialogue pairs visible to human annotators. The annotators could provide data directly on the site after viewing the videos and the scene-dialogue pairs. We converted each scene to an animated GIF to be displayed on the site. Then we asked the human annotators from *Amazon Mechanical Turk* (*AMT*) to visit the site and concretely describe each scene in one or multiple sentences following the guidelines. Total 27,328 descriptive sentences were collected from the human annotators. The average number of sentences and words in the scene description is 1.7 and 13.6. Table 1 shows the advantage of our descriptions over that of MovieQA; the descriptions are well associated with the visual stories. For the evaluation, we randomly selected 100 samples from each dataset and recruited ten persons to score on each question between 0 and 1.

Table 2: Examples and statistics by type of question

| Type | Example | Ratio |
|---|---|---|
| Action | What did *Pororo* do with the egg? | 0.20 |
| Person | Who lives in the forest? | 0.18 |
| Abstract | What is the main event of *Pororo* day? | 0.16 |
| Detail | What does the little penguin wear? | 0.15 |
| Method | How did the *Crong* introduce himself? | 0.06 |
| Reason | Why did *Pororo* take egg to home? | 0.06 |
| Location | Where is the small village situated? | 0.04 |
| Statement | What did the dinosaur say first? | 0.03 |
| Causality | What happens with *Crong* reaches the bottom of the hill? | 0.03 |
| Yes/No | Did *Pororo* took shelter with *Poby*? | 0.03 |
| Time | When did *Pororo* and his friends stop sliding downhill? | 0.02 |

**QA collection:** We recruited *AMT* workers different from the workers who participated in making the descriptive sentences. They were asked to watch the videos before creating any QA and then asked to make questions about the video stories with a correct answer and four wrong answers for each question. The descriptive sentences were not given to the annotators. Next, they gave the context for each question by localizing the question to a relevant scene-dialogue pair in the video. In other words, each question has a relevant scene-dialogue pair which contains information about the answer. After excluding QA pairs that do not follow the guidelines, e.g., vague or irrelevant ones such as “where are they?” or ‘how many trees in the videos?”, we obtained 8,913 QA pairs. The average number of QA per episode, i.e. a video, is 52.15. The average numbers of words in the question and answer is 8.6 and 7.3. Figure 2 shows the guildleines given to the *AMT* workers. Table 2 shows the examples and statistics by types of the questions.

**Dataset comparison**: We compare our dataset to other existing public video datasets in Table 3. To the best of our knowledge, PororoQA has the highest number of videoQA, as well as is the first video dataset that have a coherent storyline throughout the dataset. We plan to release the PororoQA dataset to the community as our contribution.

## 3 Deep Embedded Memory Networks

Memory networks, initially proposed for text QA, use an explicit memory component and model the relationships between the story, question, and answer [Weston *et al.*, 2015].

Table 3: Comparison of various public datasets in terms of video story analysis. N/A means that information is not available.

| Dataset | TACoS M.L [Rohrbach et al., 2014] | MPII-MD [Rohrbach et al., 2015] | LSMDC [Rohrbach et al., 2015] | M-VAD [Torabi et al., 2015] | MSR-VTT [Xu et al., 2016] | TGIF [Li et al., 2016] | MovieQA [Tapaswi et al., 2016] | PororoQA (ours) |
|---|---|---|---|---|---|---|---|---|
| # videos | 185 | 94 | 202 | 92 | 7,000 | - | 140 | 171 |
| # clips | 14,105 | 68,337 | 108,503 | 46,589 | 10,000 | 100,000 | 6,771 | 16,066 |
| # sent. | 52,593 | 68,375 | 108,470 | 46,523 | 200,000 | 125,781 | N/A | 43,394 |
| # QAs | - | - | - | - | - | - | 6,462 | 8,913 |
| Domain | Cooking | Movie | Movie | Movie | Open | Open | Movie | Cartoon |
| Coherency | X | X | X | X | X | X | X | O |

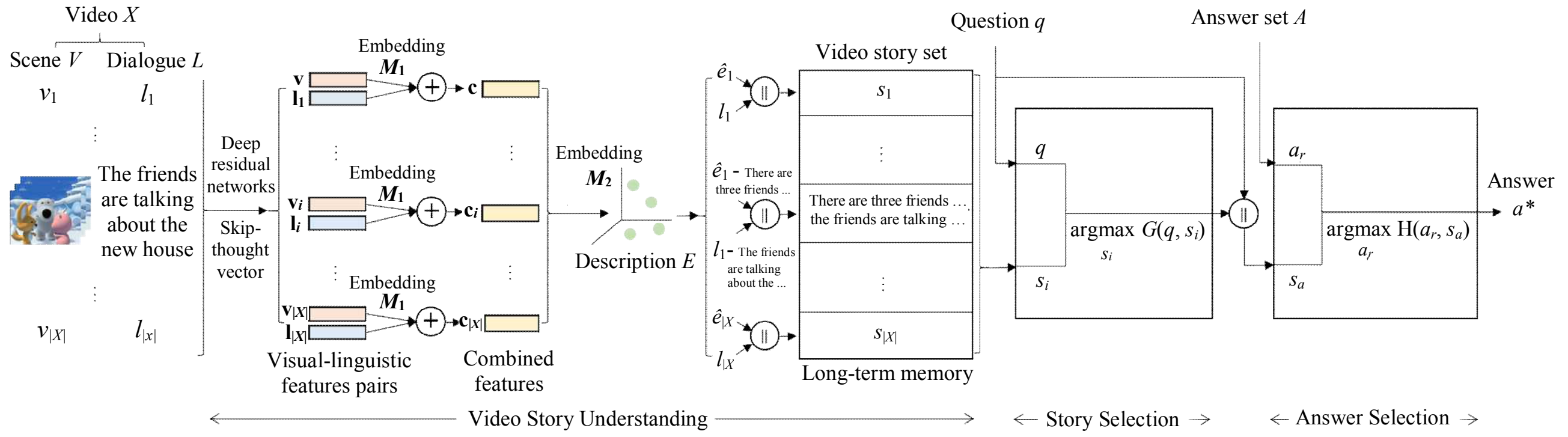

Figure 3: The structure of deep embedded memory networks for the video story QA task.

In video story QA, these ideas have to be extended such that the model understands video stories from a joint stream of two modalities, i.e scene and dialogue, and gives attention to specific pieces of evidence to answer correctly. Figure 3 shows the structure of our proposed model DEMN for video story QA. DEMN takes a video $X=\{(v_i,l_i)\}_{1\ldots|X|}$ as input, where $v_i$ is a scene (a sequence of image frames), and $l_i$ is a dialogue (a natural language sentence). A QA task passes through three modules as described below.

## 3.1 Video Story Understanding Module

The main objective of this module is to reconstruct video stories in the form of sentences from the scene-dialogue streams of the observed videos. At training time (trained independently with other modules), the module learns a scene embedding matrix $\mathbf{M_1}$ and a dialogue embedding matrix $\mathbf{M_2}$. At test time, the module transforms each scene-dialogue pair to a video story in the following way:

- Deep residual networks [Kaiming *et al.*, 2016] and an encoder-decoder deep model [Kiros *et al.*, 2015b] compute a visual-linguistic features pair ($\mathbf{v}_i$, $\mathbf{l}_i$) for an input scene-dialogue pair ($v_i$, $l_i$).
- Combined vector $\mathbf{c}_i$ is the sum of embedded representation of the scene $\mathbf{v}_i^T\mathbf{M_1}$ and representation of the corresponding dialogue $\mathbf{l}_i$, i.e. $\mathbf{c}_i = \mathbf{v}_i^T\mathbf{M_1} + \mathbf{l}_i$.
- The module retrieves the nearest description $\hat{e}_i$ to $\mathbf{c}_i$ by measuring the dot-product similarity of the embedded combined vector $\mathbf{c}_i^T\mathbf{M_2}$ and the deep representation $\hat{\mathbf{e}}_i$ of the description $\hat{e}_i$.
- We define a video story $s_i$ as a concatenation of $\hat{e}_i$ and $l_i$.

The output $S$ is a set of video stories for the input video $X$, i.e. $S=\{s_i\}_{1\ldots|X|}=\{\hat{e}_i \| l_i\}_{1\ldots|X|}$. $\|$ means concatenation. For example, $s_i$ can be 'There are three friends on the ground. The friends are talking about the new house.' Each story $s_i$ is stored in a long-term memory component, e.g. a table.

**Training:** We use the scene-dialogue-ground truth description pairs in the training dataset to learn $\mathbf{M_1}$ and $\mathbf{M_2}$. First, we train a scene embedding matrix $\mathbf{M_1}$ using a combination of hinge rank loss and dot-product score [Weston *et al.*, 2010; Frome *et al.*, 2013; Kiros *et al.*, 2015a] such that the $\mathbf{M_1}$ is trained to achieve a higher dot-product score between the embedded representation of the scene and the representation of the corresponding dialogue than the scores between non-corresponding combinations. Thus, the per training example hinge rank loss is as follows:

$$loss(v_{tr}, l_{tr}) = \sum_j \max(0, \gamma_u - \mathbf{v}_{tr}^{\mathbf{T}}\mathbf{M_1}\mathbf{l}_{tr} + \mathbf{v}_{tr}^{\mathbf{T}}\mathbf{M_1}\mathbf{l}_j) \qquad (2)$$

where $v_{tr}$ is a scene in the training video dataset and, $\mathbf{v}_{tr}$ is a vector of aggregation of image features computed from each frame of $v_{tr}$. We used the average pooling of 2,048-D sized 200-layer residual networks activations [Kaiming *et al.*, 2016]. $l_{tr}$ is a corresponding dialogue for $v_{tr}$ and, $\mathbf{l}_{tr}$ is a feature vector of $l_{tr}$ computed from 4,800-D skip-thought vectors pre-trained using *Wikipedia* and the dialogue corpus in '*Pororo*' cartoon videos [Kiros *et al.*, 2015b]. $\mathbf{l}_j$ is a feature vector of a contrastive (non-corresponding) dialogue sentence for $v_{tr}$. We use the same deep models when computing features for scenes and dialogues at the test time. $\mathbf{M_1}$ is the embedding matrix of trainable parameters pre-trained with scene-descriptive sentence pairs from MPII-MD dataset [Rohrbach *et al.*, 2015]. We use stochastic gradient descent (SGD) to train $\mathbf{M_1}$. $\gamma_u$ is a margin and fixed as 1 during training time.

After $\mathbf{M_1}$ is trained, we compute a combined vector $\mathbf{c}_{tr}$ for each pair ($v_{tr}$, $l_{tr}$) by summing the embedded scene vector $\mathbf{v}_{tr}^T\mathbf{M_1}$ and the representation of corresponding dialogue $\mathbf{l}_{tr}$. Then, we train $\mathbf{M_2}$ in the same way such that embedding of $\mathbf{c}_{tr}$, i.e. $\mathbf{c}_{tr}^T\mathbf{M_2}$, and the deep representation $\mathbf{e}_{tr}$ for the ground-truth description $e_{tr}$ of $v_{tr}$ achieves a higher dot-product score than the scores between $\mathbf{c}_{tr}$ and contrastive description vectors $\mathbf{e}_j$. Note that all denoted vectors, i.e. $\mathbf{v}_{tr}$, $\mathbf{l}_{tr}$, $\mathbf{l}_j$, $\mathbf{c}_{tr}$, $\mathbf{e}_{tr}$, $\mathbf{e}_j$ are normalized to unit length.

## 3.2 QA Modules

**Story selection module:** The key function of the module is to recall the best video story $s^*$ that contains the answer information to the question $q$. The module reads the list of the stories $S = \{s_i\}_{1\ldots|X|}$ of the input video $X$ from long-term memory and scores each story $s_i$ by matching with $q$. The highest scoring relevant story is retrieved with:

$$s^* = \arg\max_{s_i} G(q, s_i) \qquad (3)$$

where $G$ is a function that scores the match between the pair of $q$ and $s_i$. The output of the module $s_a$ is $q \parallel s^*$, which fuses the question and the relevant story. An example of $s_a$ is 'What were the friends doing on the ground? There are three friends on the ground. The friends are talking about the new house'.

**Answer selection module:** This module selects the most appropriate answer $a^*$ in the answer set $A=\{a_r\}_{1\ldots k}$. Similar to the story selection module, this module scores the match between the pair of the $s_a$ and each answer sentence $a_r$. The highest scoring answer sentence is selected with:

$$a^* = \arg\max_{a_r} H(s_a, a_r) \tag{4}$$

where $H$ is a scoring function that matches between the pair.

**Scoring function:** To handle the long sentences such as $s_i$ or $s_a$, the word level attention-based model [Tan *et al.*, 2016] is used as the scoring functions $G$ and $H$. The model builds the embeddings of two sequences of tokens $X=\{x_i\}_{1\ldots|X|}$, $Y=\{y_i\}_{1\ldots|Y|}$ and measure their closeness by cosine similarity. $X$ and $Y$ can be a video story, a question or an answer sentence. The model encodes each token of $X$, $Y$ using a bidirectional LSTM [Hochreiter *et al.*, 1997; Schuster *et al.*, 1997] and calculates the sentence vector $\mathbf{X}$ by averaging the output token vectors of the bidirectional LSTM on the $X$ side. Then the each token vector of $Y$ are multiplied by a softmax weight, which is determined by $\mathbf{X}$.

$$\mathbf{m}(t) = \tanh(\mathbf{W}_a \mathbf{h}_y(t) + \mathbf{W}_q \mathbf{X}) \tag{5}$$

$$o_t \propto \exp(\mathbf{w}_{ms}^T \mathbf{m}(t)) \tag{6}$$

$$\mathbf{h}'_y(t) = \mathbf{h}_y(t) o_t \tag{7}$$

where $\mathbf{h}_y(t)$ is the $t$-th token vector on the $Y$ side. $\mathbf{h}'_y(t)$ is the updated $t$-th token vector. $\mathbf{W}_a$, $\mathbf{W}_q$, $\mathbf{w}_{ms}$ are attention parameters. The sentence vector $\mathbf{Y}$ is calculated by averaging the updated token vectors on the $Y$ side.

**Training:** We train the QA modules in a fully supervised setting. Each question $q$ in the training data set is associated with a list of scene-dialogue pairs $\{(v_i, l_i)\}_{1\ldots|X|}$ of a video X to which the $q$ belongs and their respective judgements $\{y^s_i\}_{1\ldots|X|}$, where $y^s_i = 1$ if $(v_i, l_i)$ is correctly relevant for $q$, and $y^s_i=0$ otherwise. Also, each $q$ is associated with a list of answer sentences $\{a_r\}_{1\ldots k}$ with their judgments $\{y^a_r\}_{1\ldots k}$, where $y^a_r = 1$ if the $a_r$ is the correct answer for $q$, and $y^a_r = 0$ otherwise. In our setting, there is one relevant scene-dialogue pair and correct answer for each $q$. We considered each data instance as two triplets $(q, (v_i, l_i), y^s_i)$, $(q, a_r, y^a_r)$ and convert them to $(q, s_i, y^s_i)$ and $(s_a, a_r, y^a_r)$, where $s_i$ is $\hat{e}_i \parallel l_i$, and $s_a$ is $q \parallel \hat{e}_c \parallel l_c$. $\hat{e}_i$ means the description of $v_i$ retrieved by the video story understanding module. Subscript $c$ is an index of the correctly relevant scene-dialogue pair for $q$, i.e. $y^s_c = 1$. Training is performed with a hinge rank loss over these two triplets:

$$loss(X, E, q, A) = \sum_{s_i \neq s^*}^{|X|} \max(0, \gamma_s - G(q, s^*) + g(q, s_i)) + \sum_{a_r \neq a^*}^{k} \max(0, \gamma_a - H(s_a, a^*) + H(s_a, a_r)) \tag{8}$$

where $s^*$ is the correct relevant story for $q$, i.e. $s^* = \hat{e}_c \parallel l_c$, and $a^*$ is the correct answer sentence for $q$. $\gamma_s$ and $\gamma_a$ are margins fixed as 1 during training time.

# 4 Experimental Results

## 4.1 Experimental Setup

We split all 171 episodes of the '*Pororo*' videos into 60% training (103 episodes) / 20% validation (34 episodes) / 20% test (34 episodes). The number of QA pairs in training / validation / test are 5521 / 1955 / 1437. The evaluation methods are QA accuracy and Mean Reciprocal Rank (MRR). MRR is used to evaluate the story selection module of the models, and its value informs the average of the inverse rank of the correct relevant story among a video story set $S$.

## 4.2 Model Experiments on PororoQA

We intend to measure 1) human performance on the PororoQA task, 2) performance of existing story QA models, 3) performance comparison between the proposed model and other story QA models. The performances were evaluated for ablation experiments with all possible input combinations ($Q$: question, $L$: dialogue, $V$: scene, $E$: ground-truth descriptions). We briefly describe the human experiments, the comparative models, and our model setting.

**Human baselines:** In each experiment, six human evaluators answered all questions in the test set.

**BoW / W2V / LSTM Q+V:** These are baseline models used in the VQA challenge [Agrawal *et al*., 2015]. For video story QA task ($Q$+$L$+$V$), we extended the models by replacing the image input to the video scene input and adding an extra input (or two inputs for $L$+$E$) to the models to use linguistic sources in videos, such as dialogues or descriptions. To represent language, they used 2,000-D bag-of-word, the average pooling of 2,000-D word2vec [Mikolov *et al*., 2013], and 4,800-D skip-thoughts vectors. These linguistic features were fused with visual features calculated from the average pooling of 200-layer residual networks activations.

**Memory networks / end-to-end memory networks:** Memory networks and end-to-end memory networks [Sukhbaatar *et al.*, 2015] were initially proposed for text story QA. For the video story QA task ($Q$+$L$+$V$), these models were extended by [Tapaswi *et al.*, 2016]. They separately built two story QA models using scenes ($Q$+$V$) and dialogues ($Q$+$L$). Then fused the QA results from the last components of the models. The visual story models retrieved the descriptions $\hat{e}$ as a proxy for the scenes like our model.

**DEMN:** We evaluated the DEMN with two modes, i.e. with and without attention. We used linear neural networks as alternative scoring functions $G$ and $H$. Also, the DEMN and the (end-to-end) memory networks did not retrieve the descriptions for all ablation experiments involving $V$+$E$ but used the ground-truth descriptions instead.

**Results:** We report human performances on the PororoQA task. The first row in Table 4 shows the human performances on the experiments. Videos were important in the majority of the questions. As more information was provided, human performance was significantly improved. We found that there was more useful information in the descriptions to answer questions than the dialogues. Overall, humans achieved a high accuracy of 96.9%.

Table 4: Accuracies(%) for the PororoQA task. *Q, L, V, E* stands for question, dialogue, scene, and ground-truth description, respectively. The ablation experiments of all memory networks variants using *E* used the ground-truth descriptions $e_i$ for the scenes $v_i$, not the retrieved ones $\hat{e}_i$ (i.e. we did not use *V* if *V* and *E* are both included in the input). MRR scores are denoted in the parentheses.

| Method | *Q* | *Q+L* | *Q+V* | *Q+E* | *Q+V+E* | *Q+L+V* | *Q+L+E* | *Q+L+V+E* |
|---|---|---|---|---|---|---|---|---|
| Human | 28.2 | 68.2 | 74.3 | 70.5 | 74.6 | 96.9 | 92.3 | 96.9 |
| BoW V+Q | 32.1 | 34.6 | 34.2 | 34.6 | 34.6 | 34.4 | 34.3 | 34.2 |
| W2V V+Q | 33.3 | 34.9 | 33.8 | 34.8 | 34.0 | 34.5 | 34.6 | 34.1 |
| LSTM V+Q | **34.8** | 42.6 | 33.5 | 36.2 | 34.6 | 41.7 | 36.3 | 41.1 |
| MemN2N | 31.1 | 41.9 | 45.6 | 50.9 | | 53.7 | 56.5 | |
| MemNN | 32.1 | 43.6 (0.16) | 48.8 (0.11) | 51.6 (0.12) | | 55.3 | 58.9 | |
| DEMN w/o attn. | 31.9 | 43.4 (0.15) | 48.9 (0.11) | 51.6 (0.12) | | 61.9 (0.19) | 63.9 (0.20) | |
| DEMN | 32.0 | **47.5 (0.18)** | **49.7 (0.12)** | **54.2 (0.10)** | | **65.1 (0.21)** | **68.0 (0.26)** | |

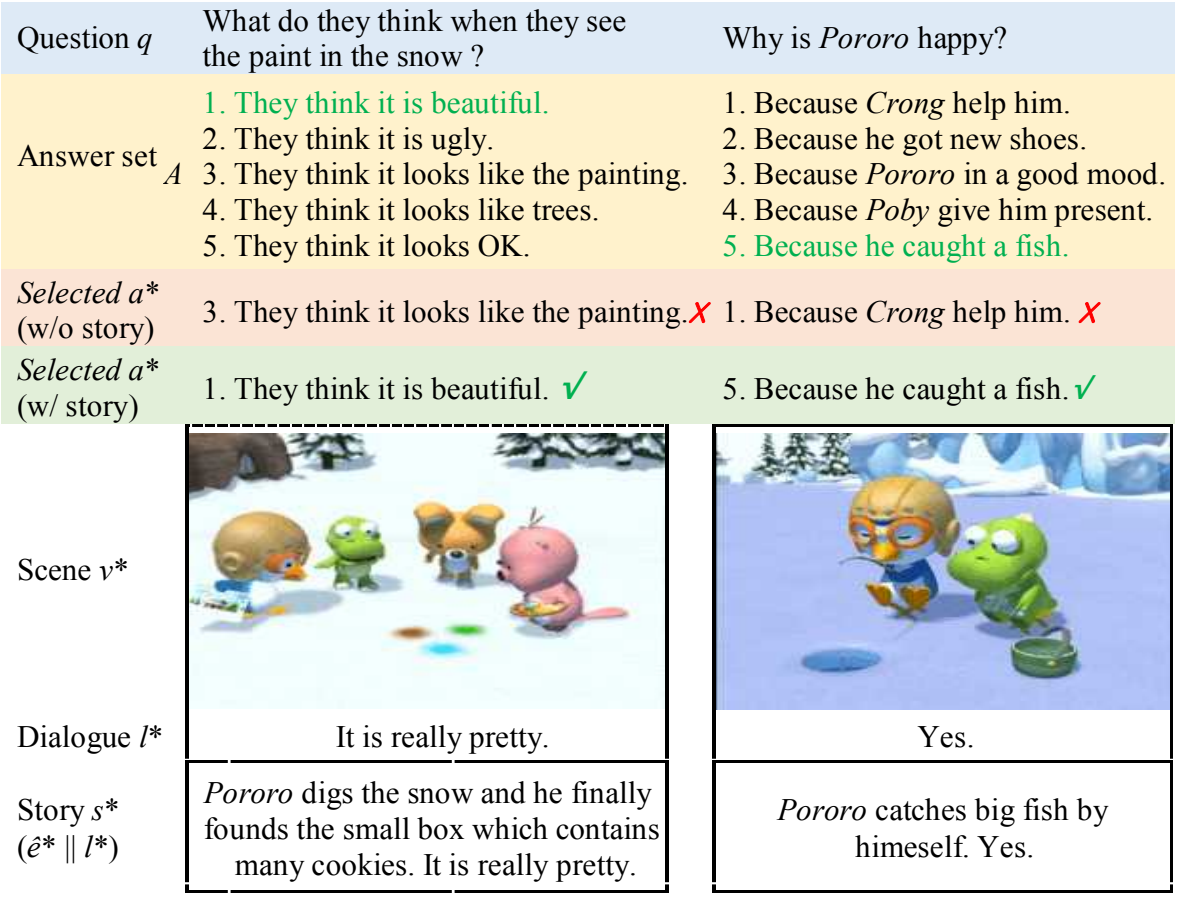

Figure 4: Qualitative results of the DEMN on the PororoQA tasks. Scene *v** and dialogue *l** correspond to the scene and dialogue that contain the story *s** selected by the model. Story *s** means the story reconstructed by the video story understanding module. We only show the first frame of the scenes.

The remaining rows in Table 4 show the performances of the QA models. Unlike images, the visualQA models failed to solve the QAs only with the visual features (*Q+V*). Also, they had difficulty using the sequences of the long dialogues and descriptions. The end-to-end memory networks do not use the supervisions between questions and relevant scene-dialogue labels, so performances were lower than the memory networks. The memory networks had similar performances to the DEMN that do not use attention when there was one story modality (*Q+L*, *Q+V*, *Q+E* or *Q+V+E*). However, when using linguistic stories and visual stories together (*Q+L+V*, *Q+L+E* or *Q+L+V+E*), DEMN was able to achieve better performances. Our combined video story reconstruction led to improved MRR scores (0.19 and 0.20) and thus QA accuracies. Using attention, DEMN showed the best performance than the other models. However, there is still room for performance improvement when comparing differences in performance with humans, or performance differences depending on whether *E* is used. Figure 4 shows the qualitative results of the DEMN on the PororoQA tasks.

Table 5: Accuracies(%) for the MovieQA task. DEMN achieved the state-of-the-art scores on the VideoQA mode. *Rand.* means the accuracy of the model is nearly 20%. SSCB is convolutional neural networks-based model [Tapaswi *et al.*, 2016].

| Method | Val | | | Test | | |
|---|---|---|---|---|---|---|
| | *Q+L* | *Q+V* | *Q+L+V* | *Q+L* | *Q+V* | *Q+L+V* |
| SSCB | 22.3 | 21.6 | 21.9 | *Rand.* | *Rand.* | *Rand.* |
| MemN2N | 38.0 | 23.1 | 34.2 | *Rand.* | *Rand.* | *Rand.* |
| DEMN | 42.4 | 39.5 | **44.7** | 28.5 | 29.4 | **30.0** |

### 4.3 Model Experiments on MovieQA Benchmark

The MovieQA benchmark dataset provides 140 movies and 6,462 multiple choices QAs. We report the accuracies of DEMN in Table 5. At the time of submission of the paper, DEMN achieved state-of-the-art results on both the validation set (44.7%) and test set (30.0%) for video QA mode. To understand the scenes, we used a description set from MPII-MD [Rohrbach *et al.*, 2015] as *E*. We assume that the reason for the relatively low performance on MovieQA is that unlike PororoQA, there are many different story structures that make optimization difficult.

## 5 Concluding Remarks

We proposed the video story QA model DEMN with the new video story QA dataset-PororoQA. PororoQA has simple, coherent story-structured videos and high-quality scene descriptions. We demonstrated the potential of our model by showing state-of-the-art performances on PororoQA and MovieQA. Our future work is to explore methods such as curriculum learning [Bengio *et al.*, 2009] that may help optimize in more complex story structures using PororoQA.

## Acknowlegdements

This work was partly supported by the ICT R&D program of MSIP/IITP. [2017-0-00162, Development of Human-care Robot Technology for Aging Society] and [2015-0-00310 -SWStarLab, Autonomous Cognitive Agent Software That Learns Real-Life with Wearable Sensors].

## References

[Abney *et al.*, 2000] Steven Abney, Michael Collins, and Amit Singhal. Answer Extraction. In *Proceedings of the 6th Applied Natural Language Processing Conference*, 296-301, 2000.

[Agrawal *et al.*, 2015] Aishwarya Agrawal, Jiasen Lu, Stanislaw Antol, Margaret Mitchell, C. Lawrence Zitnick, Dhruv Batra, and Devi Parikh. VQA: Visual Question Answering. In *Proceedings of IEEE Conference on Computer Vision and Pattern Recognition*, 2015.

[Bengio *et al.*, 2009] Yoshua Bengio, J´erˆome Louradour, Ronan Collobert, and Jason Weston, Curriculum Learning, In *Proceedings of International Conference on Machine Learning,* 2009.

[Frome *et al.*, 2013] Andrea Frome, Greg S. Corrado, Jonathon Shlens, Samy Bengio, Jeffrey Dean, Marc' Aurelio Ranzato, and Tomas Mikolov. DeViSE: A Deep Visual-Semantic Embedding Model. In *Proceedings of Advances in Neural Information Processing Systems,* 2121-2129, 2013.

[Fukui *et al.*, 2016] Akira Fukui, Dong Huk Park, Daylen Yang, Anna Rohrbach, Trevor Darrell, and Marcus Rohrbach. Multimodal Compact Bilinear Pooling for Visual Question Answering and Visual Grounding. In *Proceedings of Conference on Empirical Methods on Natural Language Processing*, 2016.

[Ha *et al*., 2015] Jung-Woo Ha, Kyung-Min Kim, and Byoung-Tak Zhang. Automated Construction of Visual-Linguistic Knowledge via Concept Learning from Cartoon Videos. In *Proceedings of AAAI Conference on Artificial Intelligence,* 2015.

[Hochreiter *et al.*, 1997] Sepp Hochreiter, and Jürgen Schmidhuber. Long short-term memory, *Neural Computation*, 9(8): 1735–1780, 1997.

[Hovy *et al.*, 2001] Eduard Hovy, Laurie Gerber, Ulf Hermjakob, Chin-Yew Lin, and Deepak Ravichandran. Toward Semantics-based Answer Pinpointing. In *Proceedings of Human Language Technology Conference*, 339-345, 2001.

[Kaiming *et al.,* 2016] Kaiming He, Xiangyu Zhang, Shaoqing Ren, and Jian Sun. Deep Residual Learning for Image Recognition. In *Proceedings of IEEE Conference on Computer Vision and Pattern Recognition*, 2016.

[Kim *et al*., 2015] Kyung-Min Kim, Chang-Jun Nan, Jung-Woo Ha, Yu-Jung Heo, and Byoung-Tak Zhang. Pororobot: A Deep Learning Robot that Plays Video Q&A Games. In *Proceedings of AAAI Fall Symposium on AI for HRI,* 2015.

[Kim *et al*., 2016] Jin-Hwa Kim, Sang-Woo Lee, Dong-Hyun Kwak, Min-Oh Heo, Jeonghee Kim, Jung-Woo Ha, and Byoung-Tak Zhang. Multimodal residual learning for visual QA. In *Proceedings of NIPS*, 2016.

[Kiros *et al*., 2015a] Ryan Kiros, Ruslan Salakhutdinov, and Richard S. Zemel. Unifying Visual-Semantic Embeddings with Multimodal Neural Language Models, In *Proceedings of Transactions of the Association for Computational Linguistics*, 2015.

[Kiros *et al.,* 2015b] Ryan Kiros, Yukun Zhu, Ruslan Salakhutdinov, and Richard S. Zemel, Antonio Torralba, Raquel Urtasun, Sanja Fidler. Skip-Thought Vectors. In *Proceedings of Advances in Neural Information Processing Systems*, 2015.

[Li *et al*., 2016] Yuncheng Li, Yale Song, Liangliang Cao, Joel Tetreault, Larry Goldberg, Alejandro Jaimes, and Jiebo Luo. TGIF: A New Dataset and Benchmark on Animated GIF Description. In *Proceedings of IEEE Conference on Computer Vision and Pattern Recognition*, 2016.

[Rohrbach *et al*., 2014] Anna Rohrbach, Marcus Rohrbach, Wei Qiu, Annemarie Friedrich, Sikandar Amin, Mykhaylo Andriluka, Manfred Pinkal, and Bernt Schiele. Coherent multi-sentence video description with variable level of detail. In *Proceedings of the German Confeence on Pattern Recognition*, 2014.

[Rohrbach *et al*., 2015] Anna Rohrbach, Marcus Rohrbach, Niket Tandon, and Bernt Schiele. A dataset for movie description. In *Proceedings of IEEE Conference on Computer Vision and Pattern Recognition*, 3202–3212, 2015.

[Schuster *et al*., 1997] Mike Schuster, and Kuldip K. Paliwal. Bidirectional Recurrent Neural Networks, *IEEE Transactions on Signal Processing*, 45(11):2673-2681, 1997.

[Sukhbaatar *et al*., 2015] Sainbayar Sukhbaatar, Arthur Szlam, Jason Weston, and Rob Fergus. End-To-End Memory Networks. In *Proceedings of Advances in Neural Information Processing Systems,* 2015.

[Tan *et al*., 2016] Ming Tan, Cicero dos Santos, Bing Xiang, and Bowen Zhou. LSTM-based Deep Learning Models for Non-factoid Answer Selection. *International Conference on Learning Representations Workshop,* 2016.

[Tapaswi *et al*., 2016] Makarand Tapaswi, Yukun Zhu, Rainer Stiefelhagen, Antonio Torralba, Raquel Urtasun, and Sanja Fidler. MovieQA: Understanding Stories in Movies through Question-Answering. In *Proceedings of IEEE Conference on Computer Vision and Pattern Recognition*, 2016.

[Torabi *et al*., 2015] Atousa Torabi, Christopher Pal, Hugo Larochelle, Aaron Courville. Using descriptive video services to create a large data source for video annotation research. *arXiv preprint arXiv:1503.01070v1,* 2015.

[Wang *et al*., 2015] Di Wang, and Eric Nyberg. A Long Short-Term Memory Model for Answer Sentence Selection in Question Answering, In *Proceedings of Association for Computational Linguistics*, 2015.

[Weston *et al*., 2010] Jason Weston, Samy Bengio, and Nicolas Usunier. Large Scale Image Annotation: Learning to Rank with Joint Word-image Embeddings. *Machine Learning*. 81(1):21-35, 2010.

[Weston *et al*., 2015] Jason Weston, Sumit Chopra, and Antoine Bordes. Memory Networks. *In Proceedings of International Conference of Learning representations,* 2015.

[Xu *et al*., 2016] Jun Xu, Tao Mei, Ting Yao, and Yong Rui. Msr-vtt: A large video description dataset for bridging video and language. In *Proceedings of IEEE Conference on Computer Vision and Pattern Recognition*, 2016.

[Yu *et al*., 2014] Lei Yu, Karl Moritz Hermann, Phil Blunsom, and Stephen Pulman. Deep Learning for Answer Sentence Selection, *arXiv preprint arXiv:1412.1632*, 2014.

[Zitnick *et al*., 2013] C. Lawrence Zitnick, and Devi Parikh. Bringing Semantics Into Focus Using Visual Abstraction. In *Proceedings of IEEE Conference on Computer Vision and Pattern Recognition*, 2013.